\documentclass[manuscript,screen,sigchi]{acmart}
\AtBeginDocument{%
  \providecommand\BibTeX{{%
    \normalfont B\kern-0.5em{\scshape i\kern-0.25em b}\kern-0.8em\TeX}}}
\setcopyright{acmcopyright}
\copyrightyear{2021}
\acmYear{2021}
\acmConference[BIOKDD '21]{BIOKDD '21: 20th International Workshop on Data Mining in Bioinformatics}{August 14--18, 2021}{Virtual}
\usepackage{bm}

\begin{document}
% \nolinenumbers
%%
%% The "title" command has an optional parameter,
%% allowing the author to define a "short title" to be used in page headers.
\title{Multi-Modal Detection of Alzheimer's Disease from Speech and Text}

%%
%% The "author" command and its associated commands are used to define
%% the authors and their affiliations.
%% Of note is the shared affiliation of the first two authors, and the
%% "authornote" and "authornotemark" commands
%% used to denote shared contribution to the research.
\author{Amish Mittal}
\authornote{Authors contributed equally to this research.}
\email{1801cs07@iitp.ac.in}
\orcid{0000-0003-2572-0322}
\affiliation{%
  \institution{Indian Institute of Technology Patna}
  \streetaddress{Bihta}
  \city{Patna}
  \state{Bihar}
  \country{India}
  \postcode{801106}
}

\author{Sourav Sahoo}
\authornotemark[1]
\email{sourav.sahoo@smail.iitm.ac.in}
\affiliation{%
  \institution{Indian Institute of Technology Madras}
  \city{Chennai}
    \state{Tamil Nadu}
  \country{India}
  \postcode{600036}
}

\author{Arnhav Datar}
\authornotemark[1]
\email{cs18b003@smail.iitm.ac.in}
\affiliation{%
  \institution{Indian Institute of Technology Madras}
  \city{Chennai}
    \state{Tamil Nadu}
  \country{India}
  \postcode{600036}
}

\author{Juned Kadiwala}
\authornotemark[1]
\email{jk555@cam.ac.uk}
\affiliation{%
 \institution{University of Cambridge}
%  \streetaddress{Rono-Hills}
 \city{Cambridge}
%  \state{Arunachal Pradesh}
 \country{United Kingdom}}

\author{Hrithwik Shalu}
\email{ae18b116@smail.iitm.ac.in}
\affiliation{%
  \institution{Indian Institute of Technology Madras}
  \city{Chennai}
    \state{Tamil Nadu}
  \country{India}
  \postcode{600036}
}

\author{Jimson Mathew}
\email{jimson@iitp.ac.in}
\affiliation{%
  \institution{Indian Institute of Technology Patna}
  \streetaddress{Bihta}
  \city{Patna}
  \state{Bihar}
  \country{India}
  \postcode{801106}
}

%%
%% By default, the full list of authors will be used in the page
%% headers. Often, this list is too long, and will overlap
%% other information printed in the page headers. This command allows
%% the author to define a more concise list
%% of authors' names for this purpose.
% \renewcommand{\shortauthors}{Trovato and Tobin, et al.}
\renewcommand{\shortauthors}{Mittal and Sahoo, et al.}
%%
%% The abstract is a short summary of the work to be presented in the
%% article.
\begin{abstract}
  Reliable detection of the prodromal stages of Alzheimer's disease~(AD) remains difficult even today because, unlike other neurocognitive impairments, there is no definitive diagnosis of AD \textit{in vivo}. In this context, existing research has shown that patients often develop language impairment even in mild AD conditions. We propose a multimodal deep learning method that utilizes speech and the corresponding transcript simultaneously to detect AD. For audio signals, the proposed audio-based network, a convolutional neural network~(CNN)-based model, predicts the diagnosis for multiple speech segments, which are combined for the final prediction. Similarly, we use contextual embedding extracted from BERT concatenated with a CNN-generated embedding for classifying the transcript. The individual predictions of the two models are then combined to make the final classification.  We also perform experiments to analyze the model performance when Automated Speech Recognition~(ASR) system generated transcripts are used instead of manual transcription and further perform an essential study of age and gender bias of our model. The proposed method achieves $85.3 \%$ 10-fold cross-validation accuracy when trained and evaluated on the Dementiabank Pitt corpus.
\end{abstract}

%%
%% The code below is generated by the tool at http://dl.acm.org/ccs.cfm.
%% Please copy and paste the code instead of the example below.
%%
\begin{CCSXML}
<ccs2012>
   <concept>
       <concept_id>10010147.10010257.10010293.10010294</concept_id>
       <concept_desc>Computing methodologies~Neural networks</concept_desc>
       <concept_significance>300</concept_significance>
       </concept>
 </ccs2012>
\end{CCSXML}

\ccsdesc[300]{Computing methodologies~Neural networks}
% \begin{CCSXML}
% <ccs2012>
%  <concept>
%   <concept_id>10010520.10010553.10010562</concept_id>
%   <concept_desc>Computer systems organization~Embedded systems</concept_desc>
%   <concept_significance>500</concept_significance>
%  </concept>
%  <concept>
%   <concept_id>10010520.10010575.10010755</concept_id>
%   <concept_desc>Computer systems organization~Redundancy</concept_desc>
%   <concept_significance>300</concept_significance>
%  </concept>
%  <concept>
%   <concept_id>10010520.10010553.10010554</concept_id>
%   <concept_desc>Computer systems organization~Robotics</concept_desc>
%   <concept_significance>100</concept_significance>
%  </concept>
%  <concept>
%   <concept_id>10003033.10003083.10003095</concept_id>
%   <concept_desc>Networks~Network reliability</concept_desc>
%   <concept_significance>100</concept_significance>
%  </concept>
% </ccs2012>
% \end{CCSXML}

% \ccsdesc[500]{Computer systems organization~Embedded systems}
% \ccsdesc[300]{Computer systems organization~Redundancy}
% \ccsdesc{Computer systems organization~Robotics}
% \ccsdesc[100]{Networks~Network reliability}

%%
%% Keywords. The author(s) should pick words that accurately describe
%% the work being presented. Separate the keywords with commas.
\keywords{Alzheimer's disease, DementiaBank, Transfer Learning, ASR, Multimodal}

%% A "teaser" image appears between the author and affiliation
%% information and the body of the document, and typically spans the
%% page.
% \begin{teaserfigure}
%   \includegraphics[width=\textwidth]{sampleteaser}
%   \caption{Seattle Mariners at Spring Training, 2010.}
%   \Description{Enjoying the baseball game from the third-base
%   seats. Ichiro Suzuki preparing to bat.}
%   \label{fig:teaser}
% \end{teaserfigure}

%%
%% This command processes the author and affiliation and title
%% information and builds the first part of the formatted document.
\maketitle

\section{Introduction}
Alzheimer's disease~(AD) is the most common underlying pathology for dementia~\cite{lam2013clinical,waldemar2007access} with existing statistics suggesting that it accounts for almost 70\% of all dementia cases~\cite{linz2018language}. AD is associated with a progressive decline in cognitive functioning such as memory, language, learning calculation, and reasoning~\cite{liu2013apolipoprotein}. Physicians, often with specialists such as neurologists, neuropsychologists, geriatricians, and geriatric psychiatrists, use various approaches and tools for diagnosis. Complementary tests include analyzing samples of cerebrospinal fluid~(CSF) taken from the brain and brain imaging tests. Such methods are invasive, expensive, and bring discomfort to the patients. Finding a noninvasive approach with ease of use with commonly used available devices is critical for early and easy detection of AD. Language impairments are usually one of the first signs of AD~\cite{klimova2015alzheimer}. In the early stages of AD, patients often develop problems linked with lexical-semantic language difficulties such as naming things or being unclear in what they say. During this phase, a patient can speak sentences that are morphologically, syntactically, and phonologically perfect; however, the sentences are often empty speech and a combination of filler words~\cite{nicholas1985empty}. Extensive studies have been conducted in detecting the patterns and markers in the speech that could indicate early-stage AD~\cite{fraser2016linguistic, altmann2001speech, hoffmann2010temporal}. 

To this end, various machine learning methods have been proposed for the detection of language impairments indicating AD from speech, text, or a combination of both~\cite{wankerl2017n,sadeghian2017speech,gosztolya2016detecting,weiner2017manual}. Most of these works use feature-based machine learning techniques to classify AD patients from the healthy control~(HC) group while a few recent approaches have approached the problem with a deep learning perspective. These are highlighted in Section~\ref{Related Work}. Traditional machine learning methods require significant domain expertise and feature engineering to extract the relevant features, and often their performance saturates with an increase in the amount of data. The use of deep learning methods are still restricted due to several reasons. Large scale expert-annotated datasets are often required for training complex deep learning models. However, curating such datasets for detecting AD based on language impairments is tedious, labour intensive, cost-prohibitive, and raises patient information confidentiality issues. 

In this work, we propose an end-to-end deep learning-based method to detect AD. We handle the data insufficiency challenge by transfer learning. Transfer learning methods have been proved to be quite successful in tasks where large scale domain-specific datasets are inaccessible~\cite{jia2018transfer,huynh2016digital}. The proposed method combines information from speech and the corresponding actual transcripts to classify people into HC or patients with AD. We also test our method on the combination of speech and transcripts generated using automatic speech recognition~(ASR) systems from the speech itself for practical purposes. Our best performing model achieves 85.3\% accuracy and 84.4\% F1 score when trained and tested on a combination of speech and actual transcripts from the DementiaBank dataset~\cite{becker1994natural}. For the combined model ASR system-generated transcripts and speech, we achieve an accuracy of 78.8\% and 78.2\% F1 score. The contributions of this work are:
\begin{enumerate}
    \item We propose a multi-modal deep learning architecture and demonstrate the efficiency of using transfer learning approaches in AD classification tasks.
    \item We investigate the viability of current ASR systems to generate transcripts for the text-based model to make the system more suited for practical purposes by eliminating the need for manual transcription of speech.
    \item Extensive experiments are performed to test for the system's bias based on gender or age.
\end{enumerate}
The rest of the paper is organized as follows: we discuss the existing works in this domain in Section~\ref{Related Work}. We describe the proposed method in Section~\ref{Method} and report the results in Section~\ref{Results}. We extrapolate our findings in Section~\ref{Discussion} and finally conclude in Section~\ref{Conclusion}.

\section{Related Work}\label{Related Work}
This section discusses some of the existing works on detecting Alzheimer's disease using machine learning techniques. Both linguistic and acoustic approaches have been used towards  AD classification. Several studies have also investigated the use of text-based features to detect AD. 

Numerous hand-crafted feature engineering-based methods have performed extremely well on the DementiaBank dataset or other such similar datasets. But, most of these datasets are relatively small in size, and hence hand-crafted methods can reasonably interpolate the data. Fraser et al.~\cite{fraser2016linguistic} used a feature selection of the best 35 features of the 370 features they extracted to get an accuracy of $87.5\%$ on the DementiaBank dataset. N-gram based approaches have also shown promising results in this field. Orimaye et al.~\cite{wankerl2017n} used manual features and N-gram-based features, while Wankerl et al.~\cite{orimaye2017} worked solely on the perplexity evaluation on the N-gram model.

With the advent of ASR technology, a few studies have relied on ASR (automatic speech recognition) and audio parameters for prediction. The study by Gosztolya et al.~\cite{gosztolya2016detecting} employed correlation-based feature selection for classification. Sadeghian et al.~\cite{sadeghian2017speech} used a customized ASR combined with feature extraction. 
% However, both papers used a limited dataset and feature engineering techniques.
%, they were able to achieve [--accuracy--] on the dataset. 
% the dataset is much smaller than ours, I don't think we should write the accuracy as it is unfair to compare. Like it was less than 100 patients for both
% - Arnhav Datar

There have been approaches where AD classification has been attempted only using audio-features. Using audio-based methods makes the system significantly less language-dependent. Chakraborty et al.~\cite{chakraborty2020identification} used an audio-only multi-class classification approach that used early and late fusion. Al-Hameed S. et al.~\cite{al2016simple} used an extensive audio-only feature set from which they extracted the best 20 features for classification to get an accuracy of $94.7\%$ on the DementiaBank dataset consisting of 477 patients. 
%However, the dataset used for both papers consisted only of English-speakers [How is ours different then?]. 

%Need to write something else here

In one of the first attempts of using deep learning based approaches in AD classification~\cite{karlekar-etal-2018-detecting}, we observe that without using any explicit acoustic or linguistic features (including POS tags), the authors achieved 84.9\% utterance segment level accuracy using a CNN-RNN based model. However, the subset of the DementiaBank dataset used by them was heavily biased towards AD-positive patients~(79.8\% AD positive, 20.2\% control group), which could have led to biased model performance. Chen et al.~\cite{chen2019} proposed an attention model that used a CNN and GRU model.  BERT-based approaches have proven to be successful on the ADReSS~\cite{luz2020alzheimers} dataset, which is a subject-independent and balanced dataset available by Dementiabank. Balagopalan et al.~\cite{balagopalan2020bert} gave a comprehensive study of both domain-based feature engineering methods and a transfer learning approach based on BERT embeddings. Pompili et al.~\cite{pompili2020inescid} used a multimodal approach to this end. They used a BERT model to get contextual embeddings, which were later fed to a bidirectional LSTM-RNN with an attention mechanism. They used a pre-trained DNN model to get acoustic embeddings from MFCC vectors, and upon fusion, they got an accuracy of $81.25\%$.
Using only sentence embeddings and a CNN-Attention network, Wang et al.~\cite{wang2021explainable} achieved an accuracy of 84.5\% on the Dementiabank dataset, while with the help of PoS features they achieved 92.2\% accuracy. They further suggested that PoS features seemed to have more value than latent features and play a higher role in detecting AD.

As the dataset~(annotated or otherwise) increases in size, feature engineering-based methods often lead to performance saturation. On the other hand, neural networks have been proved to be successful on medium-sized datasets as well as on enormous datasets like JFT-300M~\cite{sun2017revisiting} for images, AudioSet~\cite{gemmeke2017audio} for audio, and YouTube-8M~\cite{abu2016youtube} for videos. In this work, we present a deep learning based approach for both audio and textual data. So, we expect that the proposed method described in subsequent sections can scale extremely well in the future when we can access much larger domain-specific datasets.

\section{The Proposed Method}\label{Method}

\subsection{Audio-based Model}\label{Audio-based Model}
\begin{figure*}[t]
    \centering
    \includegraphics[width=0.9\textwidth]{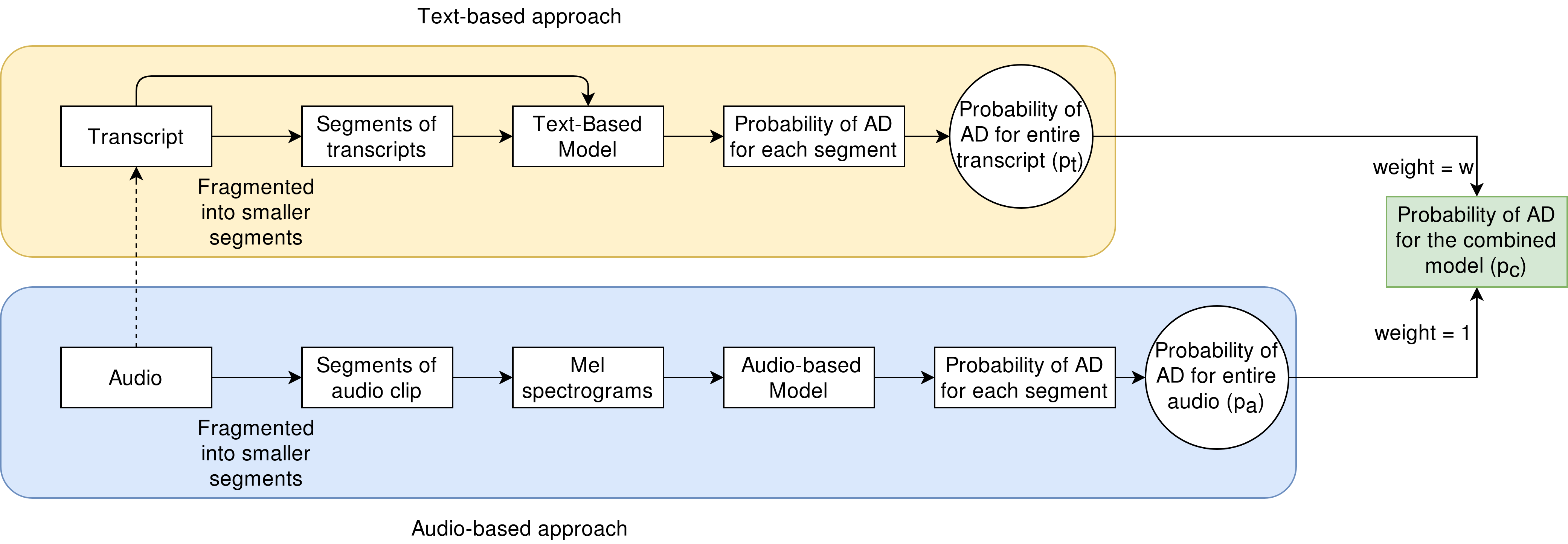}
    \caption{\textbf{Schematic diagram of the entire system.} The dotted arrow shows the scenario in which transcripts are generated from the audio clips using an ASR system.}
    \label{fig:full_system}
\end{figure*}

We process the long audio clips in a similar fashion as followed in earlier literature~\cite{sahoo2019segment}. The clips are fragmented into smaller snippets of $n$ milliseconds duration each, called \textit{segments} hereafter, which is given as input to the proposed audio-based model. An immediate advantage of such segmentation is that helps to process audios of different lengths smoothly and also allows the model to focus on the crucial features that are necessary for the classification task. In this work, we experiment with $n=960$ and $n=4960$ which are called \textit{short segments} and \textit{long segments} respectively throughout this paper. The model generates a mel spectrogram from each segment, which is passed into the pre-trained deep CNN. The 512-dimensional output from the deep CNN is fed into a single-layered neural network, predicting the probability of the segment being uttered by an AD-positive person for each segment. Finally, the arithmetic mean of the probabilities of all the segments of an audio input is calculated to classify the person into AD or HC. A schematic diagram of the method is presented in Figure~\ref{fig:full_system} and is detailed in the subsequent sections.

\subsubsection{Mel Spectrogram Generation}
The mel spectrogram generation method, as adopted from Hershey \textit{et al.}~\cite{hershey2017cnn}, is detailed as follows. For each $n$-millisecond length segment, Short-time Fourier Transform~(STFT) magnitude is computed using a 25ms length window, 10ms hop window, and Hann window smoothing function~\cite{6768513}. The spectrogram thus obtained is integrated into 64 mel-spaced frequency bins. The spectrogram is log-transformed after adding a small offset to avoid numerical instabilities. This generates log-mel spectrograms of patches of $k\times64$ bins that forms the input for the deep CNN. For short segments, we use $k=96$ and for long segments, we use $k=496$.

\subsubsection{Model Formulation}
The Google VGGish model~\cite{hershey2017cnn} is a deep CNN with an architecture similar to the VGG model~\cite{simonyan2014very} proposed for large-scale image classification. The VGGish model has been trained on AudioSet~\cite{gemmeke2017audio}, $\sim$2M human-labeled audio clips taken from YouTube videos spread over $\sim$600 audio classes. The VGGish model consists of four blocks of convolution and max-pooling layers followed by two fully connected~(FC) layers, each containing 4096 units. Finally, a 128-dimensional FC layer is present at the end, which generates the embedding vector. 

The main challenge with transfer learning is that if the target domain has limited data, such as in our case, direct fine-tuning is prone to overfitting~\cite{long2015learning}. In standard CNNs, the learned features transit from being general to task-specific along with the depth. Hence, higher layers (in this case, the FC layers) are not suitable for transfer learning via fine-tuning with a limited dataset. It has been further shown that the features learned by convolutional layers are generally task-invariant, and hence it is reasonable to preserve those layers during transfer learning~\cite{yosinski2014transferable}.

We remove the FC layers of the VGGish model and introduce a  global pooling layer~\cite{lin2013network} which produces a 512-dimensional output. We further introduce batch normalization~\cite{ioffe2015batch} in the convolutional blocks for implicit regularization and accelerate model training. The modified version of VGGish architecture is called as \textit{m-VGGish}. The m-VGGish model contains $\sim$4.8M parameters as compared to $\sim$72.2M parameters in the original VGGish model. A shallow neural network with an intermediate layer of 512 units and a two-class classification head is concatenated with the m-VGGish model, and the entire model is jointly trained in an end-to-end manner.

\subsection{Text-based Model}\label{Automatic Speech Recognition}

\begin{figure*}[t]
    \centering
    \includegraphics[width=0.69\textwidth]{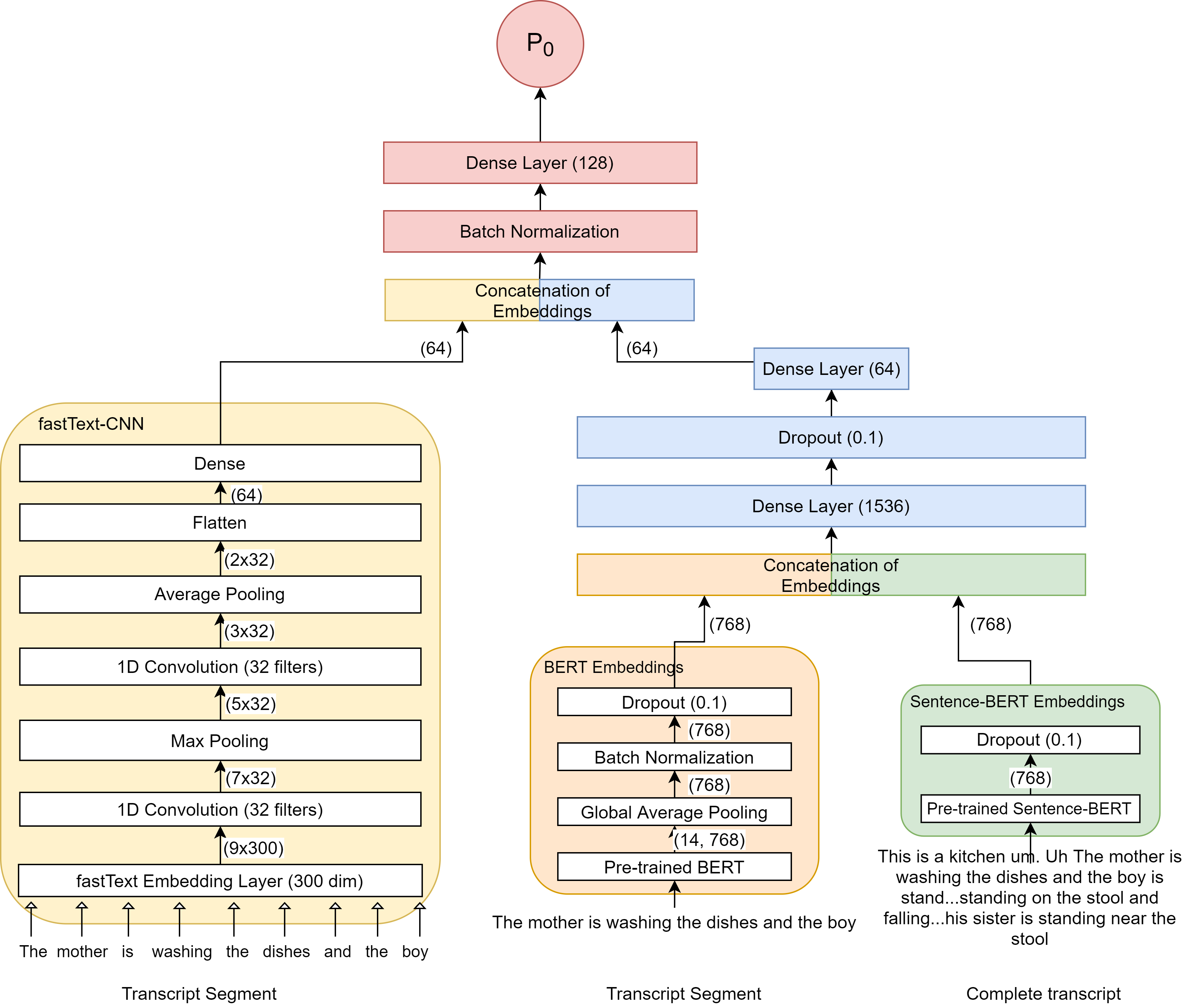}
    \caption{\textbf{Complete Architecture of the Text-based model}. Output prediction $P_0$ indicates the AD positive probability inferred from the transcript segment. Further computation is performed to calculate the patient level probability.}
    \label{fig:text_model}
\end{figure*}

The proposed text-based model comprises of the three subnetworks: 1) a CNN component, 2) a Bidirectional Encoder Representations from Transformers (BERT)~\cite{devlin2018bert} Embeddings component and 3) a SentenceBERT~\cite{reimers-2019-sentence-bert} Embeddings component. For generating the training samples, we split the entire transcript into small segments comprising seven tokens. The segments are chosen in an overlapping manner such that the last three tokens of the previous segment are the same as the first three tokens of the next segment. This is done to preserve the context of the transcript and to make sure that the language order gets represented well in each segment. All punctuation is retained in this process. So, a single training data point is a tuple consisting of a transcript segment, the corresponding transcript from which it is generated, and the target prediction of the transcript. The three different components of the text-based model emphasize on specific features of the data point as explained in the subsequent sections. The entire architecture of the text-based model is presented in Figure~\ref{fig:text_model}.

\subsubsection{fastText-CNN}
The first component of the text-based model comprises of a CNN above the fastText~\cite{grave2018learning} pre-trained word embeddings. This component of the model uses the transcript segment and patient polarity of the data point. The fastText word vectors are used due to their ability to generalize to unknown words and properly represent filler words such as \textit{uh, uhm, ohh, mhm}, which are abundantly present in the transcripts of AD-positive patients. We use CNNs to subtly capture improper speech clusters common in AD patients. These pre-trained word vectors are also fine-tuned during the training process. A fully connected layer at the end generated a 64-dimensional embedding which is concatenated with outputs from other components of the model to get the final classification probability.

\subsubsection{BERT Embeddings}
While the fastText embeddings are efficient in encoding a sentence with Out-of-Vocabulary (OOV) words, but due to the lack of current sentence context, they are often unable to represent the semantic meaning of a single or cluster of words. The second subnetwork of the model, which uses the BERT embeddings, captures the transcript segment's contextual essence. This is possible due to the encoder-decoder network that uses self-attention in the BERT model. This subnetwork is initialized with a pre-trained BERT model and is fine-tuned during model training. 

\subsubsection{SentenceBERT Embeddings}
Several works~\cite{reimers-2019-sentence-bert, xiao2018bertservice} have shown that even though BERT embeddings are highly effective for tasks such as QA~\cite{devlin2018bert} and text classification~\cite{devlin2018bert}, they often fail and instead provide poor sentence embeddings. SentenceBERT model is a modification of the pre-trained BERT that uses Siamese and Triplet network structures to derive semantically meaningful sentence embeddings~\cite{reimers-2019-sentence-bert}. As both the AD positive and HC patients tend to speak similar meaning sentences overall, i.e., describing the the Cookie-Theft Picture, it might lead to having close sentence-level BERT embeddings. So, instead of a global, low-dimensional~(768) embedding that poorly represents the whole transcript, we use SentenceBERT to capture the entire transcript's global context. 

\subsubsection{Concatenation of Embeddings}
In each training step, the SentenceBERT embeddings of the entire transcript are concatenated with the transcript segment's BERT embeddings, which are finally concatenated with the output of fastText-CNN. The concatenated output is passed through a batch normalization layer and a fully connected layer with a sigmoid activation function, predicting the probability of the patient being AD positive. It has to be noted that this architecture predicts the probability at a segment level, while the task requires prediction on a person or their transcript level. So, the probability of a particular person being AD positive is obtained by taking the arithmetic mean of the individual transcript segment's predicted probabilities.

\subsection{Late Fusion of Audio and Text-based Models}\label{combine}
Although it is advisable to learn joint representations of different modalities, in many cases such as this, it is not feasible~(or may lead to a decline in model performance). So, we adopt the late fusion strategy, which has been used in several multi-modal learning tasks~\cite{kiela2015multi, shutova2016black}. The probabilities calculated by the audio and test-based model are combined in a weighted manner, and a threshold was fixed for classifying the persons into AD and HC. Suppose $p_a$ and $p_t$ be the probability of a person being AD-positive as predicted by the audio and text-based model, respectively. Then, the combined probability is given by~\eqref{weighted}.
\begin{equation}\label{weighted}
    p_c = \frac{p_a+wp_t}{1+w}
\end{equation}{}
where $w$ is the relative weight. The three best performing combined models obtained for $w\in\{1, 1.5, 2\}$ for actual and ASR generated transcripts are presented in Table \ref{tab:all}.

\section{Experiments and Results}\label{Results}
\subsection{Dataset} \label{Dataset}
The DementiaBank corpus was collected to study communication in dementia between 1983 and 1988 at the University of Pittsburgh~\cite{becker1994natural}. It contains recordings and transcripts of English-speaking patients describing the Cookie Theft picture. The participants are categorized into dementia patient~(AD) and healthy control~(HC) groups. Of the 309 dementia samples, 235 samples are classified as probable AD, and the remaining samples are classified as other types of dementia. Our study uses only the 235 probable AD samples and 242 healthy elderly control samples.

\subsection{Experimental Setup}
The Dementia Talkbank dataset's audio files had significant background noise for generating text from the available ASR systems. The denoising was done using the minimum mean-square error log-spectral amplitude estimator method~\cite{ephraim1985speech}. We use Amazon Web Service~(AWS) Transcribe~\cite{kranz} for automated transcription. Since the dataset is primarily about the cookie theft description, the 100 most common words from the actual transcript were taken as a custom vocabulary along with a list of filler words like uh, umm, etc. 

All the experiments in this work are implemented in Keras with Tensorflow~\cite{tensorflow2015-whitepaper} backend on NVIDIA P100 GPU. The audio-based model is trained in batch-sizes of 32 using Adam optimizer~\cite{kingma2014adam}. The learning rate is set to $10^{-6}$ which is relatively less than the default value of $10^{-3}$ as we are finetuning a pre-trained model instead of training from scratch. To further prevent overfitting, early stopping~\cite{caruana2001overfitting} is used with patience value set as 30.

For the text-based model, the transcript is tokenized using the NLTK Treebank Tokenizer~\cite{10.5555/1717171}.
In the fastText-CNN subnetwork, the transcript segments are encoded using fastText pre-trained word vectors for English~\cite{fasttext} with a maximum number of tokens set equal to slightly more than the number of tokens each transcript segment contains. We use \textit{post} truncation and padding. In the BERT subnetwork, we use the max sequence length equal to double that of transcript segment length with \textit{post} padding and truncation due to the nature of the WordPiece~\cite{wu2016googles} tokenizer used in BERT. We use Hugging Face's maintained Tensorflow \textit{bert-base-uncased} model from their \textit{transformers}~\cite{Wolf2019HuggingFacesTS} library for initialization of the BERT embeddings subnetwork. SentenceBERT pre-trained model \textit{bert-base-nli-mean-tokens} is used for SentenceBERT embeddings sub-network.

\subsection{Results}
The results are evaluated using 10-fold cross-validation of the DementiaBank dataset.  All the confidence intervals were computed based on the percentiles of 1000 random resamplings~(bootstrapping) of the data. The audio-based model trained on long segments achieves an accuracy of 68.6\%~(95\% confidence interval~(CI), 65.6-71.5) and accuracy of 65.4\%~(95\% CI, 60.8-70.0) for short segments. The text-based model achieves an accuracy of 83.4\%~(95\% CI, 80.9-86.0) for actual transcripts and 75.5\%~(95\% CI, 72.3-79.2) for ASR generated transcripts. 
\begin{table}[t]
    \caption{Comparison of performances of combined model obtained by late fusion of the variants of the audio-based model~(long segments and short segments) and text-based model~(actual transcripts and ASR generated transcripts) for different weights. Acc.$=$ accuracy, F1 $=$ F1 score and $w$ is the weight for late fusion. The first row~($w = 0$) represents solely an audio-based model whereas the last row~($w= 10^{14}$) approximates a text-based model. The values in bold represent the model with best performance for each type. All the values are in percentage.}\label{tab:all}
    \centering
    \begin{tabular}{*{6}{c}}
    \toprule
    \multicolumn{1}{c}{\textbf{Type}} &
    \multicolumn{1}{c}{\textbf{\textit{w}}} &
    \multicolumn{2}{c}{\textbf{Long Segments}} & \multicolumn{2}{c}{\textbf{Short Segments}}\\
    \cmidrule(lr){3-4}\cmidrule(lr){5-6}
    &&\textbf{Acc.} & \textbf{F1}&\textbf{Acc.}& \textbf{F1}\\
    \midrule
    \midrule
    \textbf{Actual}
    &\textbf{0} & 68.6 & 69.4 & 65.4 & 66.5 \\
    &\textbf{1} & 84.7 & 83.9 & 83.4 & 82.4 \\
    &\textbf{1.5} & \textbf{85.3} & 84.4 & 82.6 & 81.1 \\
    &\textbf{2} & 84.1 & 82.6 & 82.8 & 81.2 \\
    &\bm{$10^{14}$} & 83.4 & 81.8 & 83.4 & 81.8\\
    \midrule
    \textbf{ASR}
    &\textbf{0} & 68.6 & 69.4 & 65.4 & 66.5 \\
    &\textbf{1} & 76.9 & 76.4 & 78.6 & 78.2 \\
    &\textbf{1.5} & 78.6 & 78 & 77.1 & 76.7 \\
    &\textbf{2} & \textbf{78.8} & 78.2 & 76.7 & 76.3 \\
    &\bm{$10^{14}$} & 75.5 & 74.6 & 75.5 & 74.6\\
    \bottomrule
    \end{tabular}
\end{table}{}

\begin{table*}[t]
    \caption{Comparison of our model with other related works for actual transcripts. }\label{tab:relativecomparison}
    \centering
      \centering
         \begin{tabular}{*{6}{c}}
        \toprule
        \multicolumn{1}{c}{\textbf{Related Works}} &
        % \multicolumn{1}{c}{\textbf{Year}} &
        \multicolumn{1}{c}{\textbf{Accuracy}} &
        \multicolumn{1}{c}{\textbf{Model}} &
        \multicolumn{1}{c}{\textbf{Comments}} \\
        % \multicolumn{1}{c}{\textbf{Short }}\\
        % \cmidrule(lr){3-4}\cmidrule(lr){5-6}
        % &&\textbf{Acc.} & \textbf{F1}&\textbf{Acc.}& \textbf{F1}\\
        \midrule
        \midrule
        Fraser, K. et al.~\cite{fraser2016linguistic} & 87.5 & Feature Engineering & 35 best features out of 370 \\
        Al-Hameed, S. et al.~\cite{al2016simple} &  94.7 & Feature Engineering & 20 best audio-only features \\
        Karlekar, S. et al.~\cite{karlekar-etal-2018-detecting} & 84.9 & CNN-RNN Non POS-Tagged & Utterance level prediction. Dataset biased\\
        & & &
        majority-baseline classifier achieves 79.8\% \\
        Pompili, A. et al~\cite{pompili2020inescid} & 81.25 & Multimodal Transfer Learning & ADReSS~\cite{luz2020alzheimers} test dataset accuracy \\
        Balagopalan et al.~\cite{balagopalan2020bert} & 83.3 & Transfer Learning & ADReSS~\cite{luz2020alzheimers} test dataset accuracy \\
        Wang et al.~\cite{wang2021explainable} & 92.2 & CNN with Attention & With help of PoS features (best result) \\
         Wang et al.~\cite{wang2021explainable} & 84.5 & CNN with Attention & Using universal sentence embeddings \\
        \textbf{Ours} & \textbf{85.3} & Multimodal Transfer Learning using only embeddings & - \\
        \bottomrule
        \end{tabular}

\end{table*}{}

\begin{figure}[b]
    \centering
    \includegraphics[width=\linewidth]{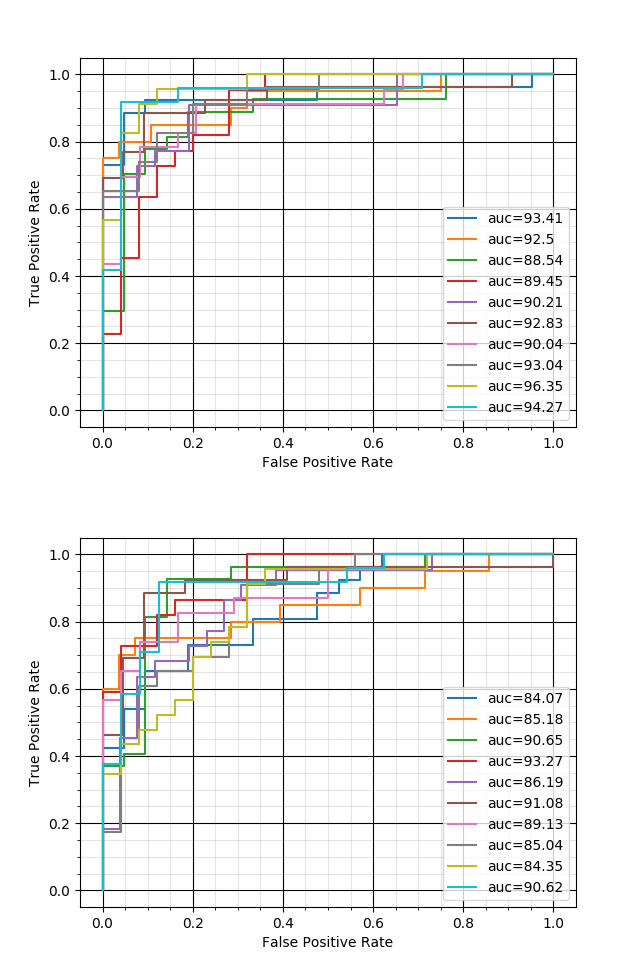}
    \caption{Receiver Operating Characteristic~(ROC) curves for the best combined model for actual~(top) and ASR generated transcripts~(bottom) for each fold of the 10-fold cross validation performed to evaluate the model performance. In both the combined models, the audio-based model use long audio segments.}\label{fig5}
\end{figure}

In Table~\ref{tab:all}, we present the accuracies and F1 scores obtained by late fusion~\cite{kiela2015multi} of the variants of the audio-based model~(long segments and short segments) and text-based model~(actual transcripts and ASR generated transcripts) for different weights. For actual transcripts, we observe that the combination of the audio-based model for long segments and the text-based model with $w=1.5$ achieves an accuracy of 85.3\%~(95\% CI, 83.0-87.8), specificity of 82.3\% and sensitivity of 89.2\% with the maximum accuracy for any fold being 93.8\%. Similarly, for ASR generated transcripts, the combination of the text-based and audio-based model for $w=2$ achieves an accuracy of 78.8\%~(95\% CI, 76.2-81.4), specificity of 78.0\% and sensitivity of 79.7\% with the maximum accuracy on any fold being 85.1\%. The best performing combined model~(model obtained by late fusion) achieves an F1 score of 84.4\%  for actual transcripts, and 78.2\% for ASR generated transcripts. The combined model's area under the curve~(AUC) score is 92.1\%~(95\% CI, 90.6-93.5) for the actual transcripts and 88.0\%~(95\% CI, 85.9-90.0) for ASR generated transcripts. The receiver operating characteristic~(ROC) curve for both the combined models is plotted in Figure~\ref{fig5}.

% \begin{figure}
% \begin{minipage}[b]{0.24\textwidth}
% \includegraphics[width=\linewidth]{ROC_new_1.png}\\
% \subcaption{ROC curve for actual transcripts}\label{fig:a0}
% \end{minipage}%
% \begin{minipage}[b]{0.24\textwidth}
% \includegraphics[width=\linewidth]{ROC_new_1.png}\\
% \subcaption{ROC curve for ASR generated transcripts}\label{fig:b0}
% \end{minipage}

% \caption{Receiver Operating Characteristic~(ROC) curves for the best combined model for actual and ASR generated transcripts for each fold of the 10-fold cross validation performed to evaluate the model performance. In both the combined models, the audio-based model use long audio segments.}\label{fig5}
% \end{figure}

\subsection{Evaluation of Model Bias}
We conduct the following analysis to check for our system's bias towards any particular age group or gender. There are primarily two reasons for this: a) there are multiple existing studies that indicate that these factors are crucial in detecting AD~\cite{vina2010women, lloret2008gender} b) the age and gender information was the only metadata available in the dataset. We compute the accuracy of our best model for different age groups and genders, which is presented in Table~\ref{tab:age} and~\ref{tab:gender} respectively. The fraction and the number of sample datapoints for each group in the dataset are highlighed. The combined model achieves an accuracy of 78.4\%~(95\% CI, 72.5-84.6) for males and 89.0\%~(95\% CI, 85.4-92.6) for females while using the actual transcripts. Similarly, for ASR generated transcripts, the combined model achieves an accuracy of 81.0\%~(95\% CI, 77.4-85.3) for females and 74.9\%~(95\% CI, 68.7-82.2) for males. 

\begin{table}[b]
    \caption{Comparison of age-wise performances of best combined model for actual transcripts and ASR generated transcripts. In both the combined models, the audio-based model use long audio segments. Acc.$=$ accuracy and F1 $=$ F1 score. The numbers in the brackets denote the actual number of participants in the category. All the values are in percentage.}\label{tab:age}
    \centering
    \begin{tabular}{*{6}{c}}
    \toprule
    \multicolumn{1}{c}{\textbf{Age Range}} &
    \multicolumn{1}{c}{\textbf{Fraction}} &
    \multicolumn{2}{c}{\textbf{Actual}} & 
    \multicolumn{2}{c}{\textbf{ASR}}\\
    \cmidrule(lr){3-4}\cmidrule(lr){5-6}
    &&\textbf{Acc.} & \textbf{F1}&\textbf{Acc.} & \textbf{F1}\\
    \midrule
    \midrule
    % \textbf{46 - 55}&0.067(32)  & 84.4          & 91.7 & 62.5 & 66.7 & 78.1          & 90.9 & 50.0   & 58.8 \\
    % \textbf{56 - 65}&0.324(154)& 85.7          & 88.8 & 80.4 & 80.4 & 81.8 & 88.9 & 71.9 & 76.7 \\
    % \textbf{66 - 75}&0.405(193)& 85.5          & 82.9 & 89.0   & 83.9 & 79.3          & 79.6 & 78.9 & 78   \\
    % \textbf{76 - 85}&0.185(88)  & 86.4 & 58.6 & 100  & 90.8 & 76.1          & 43.3 & 93.1 & 83.7 \\
    % \textbf{86 - 95}&0.021(10) & 70            & 0    & 100  & 82.4 & 50            & 0    & 100  & 66.7 \\

\textbf{46 - 55} & 0.067(32) & 84.4 & 66.7 & 78.1 & 58.8 \\
\textbf{56 - 65} & 0.324(154) & 85.7 & 80.4 & 81.8 & 76.7 \\
\textbf{66 - 75} & 0.405(193) & 85.5 & 83.9 & 79.3 & 78   \\
\textbf{76 - 85} & 0.185(88) & 86.4 & 90.8 & 76.1 & 83.7 \\
\textbf{86 - 95} & 0.021(10) & 70 & 82.4 & 50 & 66.7 \\

% \textbf{46 - 55}&	84.4&	91.7&	62.5&	66.7&	78.1&	90.9&	50&	58.8 \\
% \textbf{56 - 65}&	85.7&	88.8&	80.4&	80.4&	81.8&	88.9&	71.9&	76.7 \\
% \textbf{66 - 75}&	85.5&	82.9&	89&	    83.9&	79.3&	79.6&	78.9&	78   \\
% \textbf{76 - 85}&	86.4&	58.6&	100&	90.8&	76.1&	43.3&	93.1&	83.7 \\
% \textbf{86 - 95}&	70&	0&	100&	82.4&	50& 	0&	100&	66.&7 \\
%     \midrule
% \textbf{All}&	85.3&	82.3&	89.2&	84.4&	78.8&	78&	79.7&	78.2\\
\textbf{All}&1(477)&85.3&84.4&78.8&78.2\\

    \bottomrule
    \end{tabular}
\end{table}{}

\begin{table}[t]
    \caption{Comparison of gender-wise performances of best combined model for actual transcripts and ASR generated transcripts. In both the combined models, the audio-based model use long audio segments.  Gen. $=$ Gender, Acc.$=$ accuracy, Spec. $=$ specificity, Sens. $=$ sensitivity, F1 $=$ F1 score and Fem. $=$ Female . The numbers in the brackets denote the actual number of participants in the category. All the values are in percentage.}\label{tab:gender}
    \centering
    \begin{tabular}{*{9}{c}}
    \toprule
    \multicolumn{1}{c}{\textbf{Gen.}} &
    % \multicolumn{1}{c}{\textbf{Fraction}} &
    \multicolumn{4}{c}{\textbf{Actual}} & 
    \multicolumn{4}{c}{\textbf{ASR}}\\
    \cmidrule(lr){2-5}\cmidrule(lr){6-9}
    &\textbf{Acc.} & \textbf{Spec.}& \textbf{Sens.}& \textbf{F1}&
    \textbf{Acc.}& \textbf{Spec.}& \textbf{Sens.} & \textbf{F1}\\
    \midrule
    \midrule
    \textbf{Fem.}%&0.65(310) 
    &89.0   
    & 87.0   & 91.2 
    & 88.8  & 81.0   
    & 80.3 & 81.7 
    & 80.9  \\
    \textbf{Male}% &0.35(167)
    &78.4 
    & 74.8 & 84.4 
    & 75.0  & 74.9
    & 74.2 & 75.7 
    & 72.7 \\ 
    \midrule
    \textbf{All}%&1(477)
    &85.3
    &82.3&89.2
    &84.4&78.8
    &78.0&79.7
    &78.2\\
    \bottomrule
    \end{tabular}
\end{table}{}

\section{Discussion}\label{Discussion}
For the audio-based model, the reduction in accuracy and spreading of the confidence interval, indicating an increase in uncertainty, when we use short segments instead of long segments is expected because the latter can capture the various speech errors such as incoherent phrases, semantic and graphemic paraphasia~\cite{taler2008language}. The text-based model trained on ASR generated transcripts suffers a 7.8\% decrease in accuracy compared to the model trained on actual transcripts. There are several reasons for this observation. Apart from low Word Error Rate (WER), the ASR models tuned for this task should be able to capture out speech filler words (\textit{mhm, uh, oh, hm, etc.}) and the breaks in speech flow using explicit tokens. However, most of the existing ASR systems often do not retain punctuation marks in the generated text's appropriate locations. Secondly, modern ASR systems are data-driven, and these systems are often trained on healthy professional speakers' data. So, these systems often automatically correct the speech errors~(spellings, broken words) of the AD patients, which are crucial for classification. This observation shows the need for research in such ASR solutions by the community before these diagnostic methods can be brought to scale for the general public. 

We compare the results of our model with some existing methods in Table~\ref{tab:relativecomparison}. We compare the accuracy only as it is the common metric reported by all the methods. We observe that our method performs better in comparison to most of the existing deep learning method. The traditional methods perform better compared to our model on the given dataset. The primary reason for the same is the small quantity and low diversity of the dataset which provides the scope for feature engineering methods. However, we speculate that once the quantity and diversity of available data increases in the future, the proposed method will be able to scale appropriately whereas the performance of the hand-engineered models will saturate. The main difference between the proposed method and the existing learning-based multi-modal approach~\cite{pompili2020inescid} is the design of the acoustic system. Even though the authors use a DNN for final prediction, the input for the same is low-dimensional~(30) MFCC features as compared to the $496\times 64$-dimensional mel-spectrograms used in our case. Secondly, the VGGish model is pre-trained on a significantly larger and varied dataset. These differences lead to a superior performance of the proposed audio-only model~(68.6\%) vs 54.2\% in~\cite{pompili2020inescid}.

We pick up the top 5 segments of the transcript according to the predicted probabilities to help get a deeper insight into the changing linguistic and cognitive patterns in Alzheimer's patients, which can be viewed by linguists and neurologists for further analysis.  This highlighting method decreases the dependence on the black-box approach of the neural network and provides a way of manual validation of the model prediction by a suitable expert diagnosing the disease. An example of such highlighting paradigm is available in Table \ref{highlight-demo} for manual transcripts and Table \ref{highlight-demo-asr} for ASR generated transcripts.

% This is also viewable to a doctor diagnosing a patient for Alzheimer's disease using this approach, and can used as a further step in the diagnosis process.
\begin{table}[!tbh]
    \caption{Highlighting of possible AD indicators in sample actual transcripts. Predictions for the highlighted segment were more than 0.96, while the rest had a significantly lower values, with values being close to (0.02-0.2) in AD-positive examples and 0.05 in the AD-negative examples. All three examples were classified correctly. Best viewed in colour.}\label{highlight-demo}
    \centering
    \begin{tabular}{p{2cm}p{6cm}}
    \toprule
       \textbf{True AD-positive}  & okay he's falling \textcolor{red}{off a chair.she's uh running} the water over. can't see \textcolor{red}{anything else.no. okay.she 's she's step} in the water.no.\\\midrule
           \textbf{True AD-positive} & \textcolor{red}{mhm. there's a young boy uh going} in a cookie jar.
    and there's a  lit a girl  young girl.
    and I'm saying he's a boy because you can  hard it's hardly  hard to tell \textcolor{red}{anymore.
    uh and he's  he's in the c t cookie} jar.
    and there's a s stool that he is on and it already is starting to fall over.
    and so is the \textcolor{red}{water in the sink uh is ev overflowing in the sink. 
    hm I  I don't} know about \textcolor{red}{the  this hickey here I  whether that's more than what I said.  
    uh like it uh the wife or g Imean  uh the  the mother} is near the girl.
    and \textcolor{red}{she's  uh w uh h she has  uh has  /. 
    oh uh I  I can't think of the  ... 
    she has uh the} she's trying to wipe  uh wipe dishes.
    \textcolor{red}{oh a and stop the water from going out.}
    \\\midrule
    \textbf{True AD-negative}&well the kids are in the kitchen with their mother uh uh taking cookies out of the cookie jar.a boy's handing it to the girl.and the boy's \textcolor{red}{uh on a on a uh stool and and he's} tripping over.he's gonna fall on the floor.the mother's standing there doing the dishes.she's washing the dishes looking out the open window.and the water's running \textcolor{red}{down over the sink on on the} floor getting her feet wet.and there are she's drying a dish.and there are a couple of dishes sitting on the k kitchen counter .and looking out the window uh it's probably in the spring or summer of the year. \\
    \bottomrule
    \end{tabular}
    \end{table}

\begin{table}[!tbh]
    \caption{Highlighting of possible AD indicators in sample ASR generated transcripts. Predictions for the highlighted segment were more than 0.8, while the rest had a significantly lower values, with major values ranging from (0.4-0.75) in AD-positive examples and from (0.1-0.6) in the AD-negative examples. The ASR solution was not able to capture the interviewee's voice in the 1st example. We include this as a case where the model predicted AD-positive due to interruptions by the interviewer. Best viewed in colour.}\label{highlight-demo-asr}
    \centering
        \begin{tabular}{p{2cm}p{6cm}}
    \toprule
         \textbf{True AD-positive} & okay. And there's the picture. Tell reality act okay here. Anything? Yeah, running water over and I in the anymore. Action. Oh, okay. Carney. Okay. Anything else? Okay. Okay, Mr Stearns, that \\
         \midrule
         \textbf{True AD-positive}&you see Going on in the picture. Tell \textcolor{red}{me all the action. Okay. There's} a young boy, uh, going in a cookie jar. And there's a girl, young girl, and I'm saying he's a boy because you cannot already hard two tell anymore. Uh, he's easing two cookie cookie jar, and there's a stool that he is on, and it already is starting to fall over. And so is the water in the sink is overflowing in the \textcolor{red}{sink. Okay. Anything else? Hmm? I don't know} about the sticky here, whether that's more than what I said, \textcolor{red}{like, two go the wife} or I mean, the mother is near the girl, and \textcolor{red}{she's ah, uh, where he} has, \textcolor{red}{uh uh, I can't} think of the She has, uh, good. She's trying to wipe white dishes. Oh, and stopped a water from going up. \textcolor{red}{Okay, okay. Good}\\
         \midrule
         \textbf{True AD-negative}&\textcolor{red}{oh, show your picture. Neither picture.} There's things going on. Some action taking place I want you to do is look at the picture and just tell me anything you see going on. Well, kids air in the kitchen with her mother, taking cookies out of the cookie jar, boys handing it to the girl. The boys on stool when he's tripping over his going falling, uh, floor mother standing, They're doing the dishes. She's washing the dishes, looking out the open window. Ah, know waters running down over the sink, gon on the floor, getting her feet wet. And our she's drying a dish and a couple of dishes, sitting on the kitchen counter and looking out that window. It's probably in the spring or summer of the year\textcolor{red}{, okay?}\\
         \bottomrule
    \end{tabular}
    \end{table}
The combined model performs better than the individual text-based and audio-based model in all cases except when the audio-based model using short segments and text-based model using the actual transcripts are combined~(see Table~\ref{tab:all}). We observe that the combined model is not biased to any particular age group except the 86-95 age group, where the model accuracy is much lower than the overall accuracy~(70.0\% vs. 85.3\% for actual transcripts and 50.0\% vs. 78.8\% for ASR generated transcripts). The lower accuracy for this age group can be attributed to a sample size of only ten persons~($\sim$2.1\% of the entire dataset). On observing the dataset, we find that speech impairment in this age group was common in both control and AD-positive patients, which could be attributed to other causes. We also find that the combined model is not biased towards gender as both specificity and sensitivity have similar increasing trends between males and females. The lower detection accuracy in males than females can possibly be attributed to fewer male participants (167) than female participants (310) and also to the fact that the incidence of AD in females is more than males~\cite{vina2010women}. 

\section{Conclusion}\label{Conclusion}
In this paper, we propose an end-to-end multi-modal deep learning based method for detecting AD from speech and text. We combine two transfer-learning based approaches based on text and audio and observe its performance on the Dementia Talkbank database. Our results show that an end-to-end deep learning based model can be used effectively for the AD classification task. We intend to propose the above methodology, in its current form, to act as a widely available screening method for early detection of AD. In the future, we intend to design a model that uses an early fusion of speech, text, and other probable modes of information such as video to obtain a system with better performance. 

% \section{Figures}

% The ``\verb|figure|'' environment should be used for figures. One or
% more images can be placed within a figure. If your figure contains
% third-party material, you must clearly identify it as such, as shown
% in the example below.
% \begin{figure}[h]
%   \centering
%   \includegraphics[width=\linewidth]{sample-franklin}
%   \caption{1907 Franklin Model D roadster. Photograph by Harris \&
%     Ewing, Inc. [Public domain], via Wikimedia
%     Commons. (\url{https://goo.gl/VLCRBB}).}
%   \Description{A woman and a girl in white dresses sit in an open car.}
% \end{figure}

% Your figures should contain a caption which describes the figure to
% the reader.

% Figure captions are placed {\itshape below} the figure.

% Every figure should also have a figure description unless it is purely
% decorative. These descriptions convey what’s in the image to someone
% who cannot see it. They are also used by search engine crawlers for
% indexing images, and when images cannot be loaded.

%%
%% The acknowledgments section is defined using the "acks" environment
%% (and NOT an unnumbered section). This ensures the proper
%% identification of the section in the article metadata, and the
%% consistent spelling of the heading.
% \begin{acks}
% To Robert, for the bagels and explaining CMYK and color spaces.
% \end{acks}

%%
%% The next two lines define the bibliography style to be used, and
%% the bibliography file.
\bibliographystyle{ACM-Reference-Format}
\bibliography{sample-base}

%%
%% If your work has an appendix, this is the place to put it.
% \appendix

\end{document}